\documentclass[11pt]{article}
\usepackage{coling2018}
\usepackage{times}
\usepackage{url}
\usepackage{latexsym}
\usepackage{comment}
\usepackage{colortbl}
\usepackage{subcaption}
\usepackage{float}
\usepackage{todonotes}

\usepackage{rotating}
\usepackage{array}
\usepackage{colortbl}
\usepackage{pifont}
\usepackage{adjustbox}

\title{Personality Trait Detection\\Using Bagged SVM over BERT Word Embedding Ensembles}

\author{Amirmohammad Kazameini$^1$, Samin Fatehi$^1$, Yash Mehta$^2$, Sauleh Eetemadi$^1$, Erik Cambria$^3$  \\
 $^1$School of Computer Engineering, Iran University of Science and Technology, Iran\\
 $^2$Gatsby Computational Neuroscience Unit, University College London, UK\\
 $^3$School of Computer Science and Engineering, Nanyang Technological University, Singapore\\
\texttt{\{a\_kazemeini,sa\_fatehi\}@comp.iust.ac.ir}, \texttt{y.mehta@ucl.ac.uk},\\\texttt{sauleh@iust.ac.ir}, \texttt{cambria@ntu.edu.sg}}
\date{}

\begin{document}
\maketitle
\begin{abstract}
Recently, the automatic prediction of personality traits has received increasing attention and has emerged as a hot topic within the field of affective computing. In this work, we present a novel deep learning-based approach for automated personality detection from text. We leverage state of the art advances in natural language understanding, namely the BERT language model to extract contextualized word embeddings from textual data for automated author personality detection. Our primary goal is to develop a computationally efficient, high performance personality prediction model which can be easily used by a large number of people without access to huge computation resources. Our extensive experiments with this ideology in mind, led us to develop a novel model which feeds contextualized embeddings along with psycholinguistic features  to a Bagged-SVM classifier for personality trait prediction. Our model outperforms the previous state of the art by 1.04\% and, at the same time is significantly more computationally efficient to train. We report our results on the famous gold standard Essays dataset for personality detection.
\end{abstract}

\section{Introduction and Related Work}
An individual’s personality has a great impact on their lives, affecting their life choices, well-being, health and even preferences and desires. Hence, there is a huge interest to develop models which can automatically identify an individual’s personality with important practical applications such as in recommendation systems~\cite{yin2018network}, job screening~\cite{liem2018psychology}, social network analysis~\cite{maria2014personality}, etc. Our model makes binary predictions of the author’s personality based on the famous Big-Five~\cite{digman1990personality} personality measure, which are the following five traits: Extraversion (EXT), Neuroticism (NEU), Agreeableness (AGR), Conscientiousness (CON) and Openness (OPN).

Common author personality detection techniques usually involve extracting psycholinguistic features from text, such as Linguistic Inquiry and Word Count (LIWC)~\cite{pennebaker2001linguistic}, Mairesse features~\cite{mairesse2007using}, and SenticNet~\cite{camnt5}, which are then fed into traditional machine learning classifiers such as support vector machine (SVM)~\cite{hearst1998support}, Na\"ive Bayes, etc. More recent work leverage deep learning and make use of pre-trained word embeddings like Word2Vec~\cite{mikolov2013efficient} and Glove~\cite{pennington2014glove}. Recently, ~\cite{mehta2020personality} reviewed the latest advances in deep learning-based automated personality detection from the viewpoint of different input modalities along with recent techniques for effective multimodal personality prediction. 

The previous state of the art~\cite{majumder2017deep} on the Essays dataset also make use of a deep learning based approach with a convolutional network on top of word embeddings extracted from Word2Vec. They also incorporate other inputs such as the Mairesse features, word count, average sentence length, etc. for their final prediction. Their model outperformed the previous best ~\cite{mohammad2015using} by 0.55\%, whereas our model outperforms~\cite{majumder2017deep} by 1.04\% and at the same time being significantly more computationally efficient to train.

\begin{figure*}[t]
\centering
\includegraphics[width=1\textwidth]{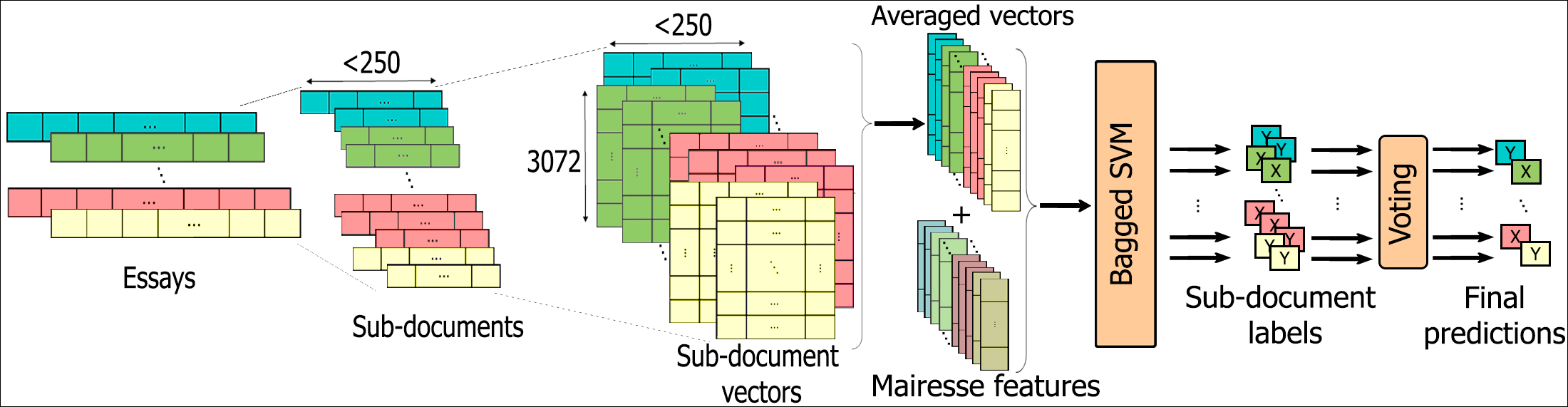} 
\caption{An overview of our deep learning-based Bagged-SVM model for automated personality detection}
\label{figure proposed method}
\end{figure*}

\section{Proposed Method}
Each of our inputs is an essay with a mean size of around 650 words. The maximum number of tokens BERT can process at a time is 512. Hence, to extract maximum information from the input, we break the essay into multiple chunks (sub-documents), with the maximum length of a chunk being 200 tokens. All these sub-documents of a particular essay are then annotated with the same personality label as that essay. We experimented with various methods of pre-processing the essays text before feeding it to the BERT tokenizer and use the best performing one.

We split the text into a sequence of sentences at the period and question mark characters and remove all characters other than ASCII letters, digits, quotations and exclamation marks. We expand all shortenings (e.g., ``you're" becomes ``you are") which increases the maximum length of a sub-document from 200 to 250 tokens. After this initial pre-processing step, the sub-documents are then fed into the pre-trained BERT\textsubscript{BASE} model. For each layer of BERT, we average the contextual token representations of that layer. Then, we concatenate the last four layer representations and concatenate this with the corresponding 84 Mairesse features for the essay. This is then considered as the feature vector for the document, which is of the dimension ${\rm I\!R}^{3156}$.

In the classification phase, we feed the document feature vector to a SVM which predicts a binary label corresponding to a particular personality trait. To enhance the performance further, we use ten SVM classifiers to perform the prediction in parallel like the bagging classifier~\cite{breiman1996bagging}. The estimator trains on all the features on the total number of the sub-document stack with replacement and the final predicted model for a document is done by majority voting. 

\section{Evaluation}
Our model achieves a 1.04\% increase in performance in comparison to the previous state of the art along with being significantly more computationally efficient. To put it in perspective, ~\cite{majumder2017deep} train their model on an Intel Core i7-4720 HQ CPU with the optimal configuration and it takes about 50 hours to complete. Our fine-tuning model only takes about 7 minutes to train. Table~\ref{table 1} gives a comparison of our model, BB-SVM, with others. We modify various parts of the BB-SVM model and discuss their effect on performance in the following section.

\begin{table}[h]
\begin{minipage}[t]{.48\linewidth}
\setlength{\tabcolsep}{3.4pt}
\renewcommand{\arraystretch}{1.3}
\scriptsize
\begin{tabular}{l|ccccc|c}
\cline{2-6}
    &\multicolumn{5}{|c|}{\textbf{Personality Traits}}&\\
    \hline
    \multicolumn{1}{|l|}{\textbf{Model Name}} & \textbf{EXT} & \textbf{NEU} & \textbf{AGR} & \textbf{CON} & \textbf{OPN} & \multicolumn{1}{|c|}{\textbf{Average}}\\ \hline
     \multicolumn{1}{|l|}{Majority Baseline} & 51.72 & 50.2 & 53.10 & 50.79 & 51.52 & \multicolumn{1}{|c|}{51.43}\\ \hline
    \multicolumn{1}{|l|}{Mairesse} & 55.13 & 58.9 & 55.35	& 55.28	& 59.57 & \multicolumn{1}{|c|}{56.84}\\ \hline
    \multicolumn{1}{|l|}{Previous state of the art} & 58.09 & 57.33 & \textbf{56.71} & 56.71 & 61.13 & \multicolumn{1}{|c|}{57.99} \\ \hline
    \multicolumn{1}{|l|}{BB-SVM} & \textbf{59.30} & \textbf{59.39} & 56.52 & \textbf{57.84} & \textbf{62.09} & \multicolumn{1}{|c|}{\textbf{59.03\textsuperscript{\ding{61}}}}\\
    \hline
\end{tabular}%
\caption{A comparison of the performance of our model (BB-SVM) with others (\ding{61}: Statistically significant at p $\le$ 0.05)}
\label{table 1}
\end{minipage}%
\hfill
\begin{minipage}[t]{.48\linewidth}
\scriptsize
\centering
\setlength{\tabcolsep}{2pt}
\renewcommand{\arraystretch}{1}
\begin{tabular}{|c|c|c|c|c|c|}
\hline
\rule{0pt}{16pt}\textbf{\shortstack{Model \\ Id}} & \textbf{\shortstack{Word \\ Embedding}} & \textbf{\shortstack{Sentence \\ Feature \\ Extraction}} & \textbf{\shortstack{Document \\ Feature \\ Extraction}} & \textbf{Classifier} & \textbf{\shortstack{Average \\ Accuracy}} \\ \hline
M8 & W2V & - & Mean & Bagging-SVM & 57.38 \\ \hline
\tiny{BB-SVM} & \shortstack{BERT \\ \tiny{(4 last layers)}} & - & Mean & Bagging-SVM & 59.03 \\ \hline
\end{tabular}%
\caption{A comparison of average accuracy of all the 5 traits with different word embeddings}
\label{table word embedding}
\end{minipage}%
\end{table}

\subsubsection{Word Embedding}
We compare the model performance with context-independent word embeddings such as Word2Vec. Table~\ref{table word embedding} shows a comparison of the results. Also, results reported by~\cite{devlin2018bert} suggest that concatenating the last four layers of BERT gives the best representation for a word. The comparison of BB-SVM results with inputs of different BERT layers is shown in Figure~\ref{fig3}.

\begin{minipage}{0.92\textwidth}
\begin{minipage}[t]{0.5\textwidth}\vspace{0pt}%
\centering
\includegraphics[width=\linewidth]{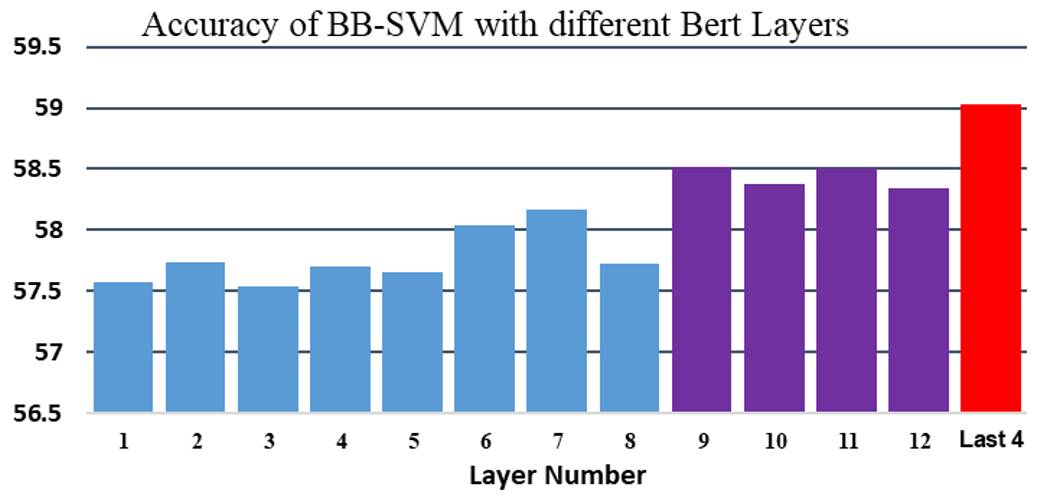} 
\captionof{figure}{Accuracy of BB-SVM model with different BERT layer inputs compared to the accuracy of this model with the concatenation of the last four layers}
\label{fig3}
\end{minipage}%
\hfill
\begin{minipage}[t]{0.48\textwidth}\vspace{15pt}%
\scriptsize
\centering
\setlength\tabcolsep{2pt}
\renewcommand{\arraystretch}{1.3}
\begin{tabular}{|c|c|c|c|c|c|}
\hline
\rule{0pt}{16pt}\textbf{\shortstack{Model \\ Id}} & \textbf{\shortstack{Word \\ Embedding}} & \textbf{\shortstack{Sentence \\ Feature \\ Extraction}} & \textbf{\shortstack{Document \\ Feature \\ Extraction}} & \textbf{Classifier} & \textbf{\shortstack{Average \\ Accuracy}} \\ \hline
\tiny{DocBERT} & BERT & - & - & MLP & 57.11 \\ \hline
M11 & \shortstack{BERT \\ \tiny{(layer 11)}} & Mean & CNN+Max & MLP & 57.42 \\ \hline
M12 & \shortstack{BERT \\ \tiny{(layer 11)}} & Mean & CNN+GRU & MLP & 57.42 \\ \hline
M3 & \shortstack{BERT \\ \tiny{(layer 11)}} & - & Mean & SVM & 58.49 \\ \hline
M14 & \shortstack{BERT \\ \tiny{(layer 11)}} & - & Mean & Bagging-SVM & 58.51 \\ \hline
\end{tabular}
\captionof{table}{A comparison of average accuracy of all the 5 traits with different base classifiers}
\label{table classifier}
\end{minipage}%
\end{minipage}%

\subsubsection{Fine-tuning Network and Feature Extraction}
In the classification phase, we experimented with a SVM and a multi-layer perceptron for making the final personality trait predictions. We found that using a SVM results in better performance. We also experiment with feeding sub-document features to DocBERT~\cite{adhikari2019docbert}, followed by averaging sub-document predictions to obtain the document's prediction. However, this did not improve the results. Table~\ref{table classifier} shows a comparison of the results. We train the model by applying Bagging (using ten simultaneous SVM classifiers) and in line with previous studies~\cite{kim2002support}, Bagging improved the classification accuracy for the task of personality detection as well (table~\ref{table bagging}).

We also tried 2 different ways to extract the document features. In the traditional approach, the document features are directly constructed from word features. A different approach is to first construct sentence features using the word features and then construct the document features from these sentence features. Table~\ref{table feature extraction} shows a comparison of the results.

\begin{table}[h]
\begin{minipage}[t]{.48\linewidth}
\scriptsize
\centering
\setlength\tabcolsep{2pt}
\begin{tabular}{|c|c|c|c|c|c|}
\hline
\rule{0pt}{16pt}\textbf{\shortstack{Model \\ Id}} & \textbf{\shortstack{Word \\ Embedding}} & \textbf{\shortstack{Sentence \\ Feature \\ Extraction}} & \textbf{\shortstack{Document \\ Feature \\ Extraction}} & \textbf{Classifier} & \textbf{\shortstack{Average \\ Accuracy}} \\ \hline
M13 & \shortstack{BERT \\ \tiny{(4 last layers)}} & - & Mean & SVM & 58.76 \\ \hline
\tiny{BB-SVM} & \shortstack{BERT \\ \tiny{(4 last layers)}} & - & Mean & Bagging-SVM & 59.03 \\ \hline
\end{tabular}
\caption{A comparison of average accuracy of all the 5 traits with and without applying bagging}
\label{table bagging}
\end{minipage}%
\hfill
\begin{minipage}[t]{.48\linewidth}
\scriptsize
\centering
\setlength\tabcolsep{2pt}
\renewcommand{\arraystretch}{1}
\begin{tabular}{|c|c|c|c|c|c|}
\hline
\rule{0pt}{16pt}\textbf{\shortstack{Model \\ Id}} & \textbf{\shortstack{Word \\ Embedding}} & \textbf{\shortstack{Sentence \\ Feature \\ Extraction}} & \textbf{\shortstack{Document \\ Feature \\ Extraction}} & \textbf{Classifier} & \textbf{\shortstack{Average \\ Accuracy}} \\ \hline
M9 & \shortstack{BERT \\ \tiny{(4 last layers)}} & Mean & Mean & Bagging-SVM & 57.91 \\ \hline
\tiny{BB-SVM} & \shortstack{BERT \\ \tiny{(4 last layers)}} & - & Mean & Bagging-SVM & 59.03 \\ \hline
\end{tabular}
\caption{A comparison of average accuracy of all the 5 traits with different features extraction methods}
\label{table feature extraction}
\end{minipage}%
\end{table}

\section{Conclusion and Future Work}
In this paper, we presented a computationally efficient deep learning-based model which outperformed the state of the art on the famous stream of consciousness Essays dataset. We hope that our model can be useful for research teams which do not have access to large computational resources. We believe a promising direction of future research would be to make more interpretable deep learning models which can provide valuable insights into the main psychological features driving these predictions and in turn also help advance psychological studies. Currently, the availability of quality personality datasets is quite limited. If an individual's personality can predicted with a little more reliability, there is scope for integrating automated personality detection in almost all human-machine interaction agents such as voice assistants, robots, cars, etc. 

\newpage

\bibliographystyle{acl}
\bibliography{WiNLP}

\begin{thebibliography}{}

\bibitem[\protect\citename{Adhikari \bgroup et al.\egroup
  }2019]{adhikari2019docbert}
Ashutosh Adhikari, Achyudh Ram, Raphael Tang, and Jimmy Lin.
\newblock 2019.
\newblock Docbert: Bert for document classification.
\newblock {\em arXiv preprint arXiv:1904.08398}.

\bibitem[\protect\citename{Breiman}1996]{breiman1996bagging}
Leo Breiman.
\newblock 1996.
\newblock Bagging predictors.
\newblock {\em Machine learning}, 24(2):123--140.

\bibitem[\protect\citename{Cambria \bgroup et al.\egroup }2018]{camnt5}
Erik Cambria, Soujanya Poria, Devamanyu Hazarika, and Kenneth Kwok.
\newblock 2018.
\newblock {SenticNet} 5: Discovering conceptual primitives for sentiment
  analysis by means of context embeddings.
\newblock In {\em {AAAI}}, pages 1795--1802.

\bibitem[\protect\citename{Devlin \bgroup et al.\egroup }2018]{devlin2018bert}
Jacob Devlin, Ming-Wei Chang, Kenton Lee, and Kristina Toutanova.
\newblock 2018.
\newblock Bert: Pre-training of deep bidirectional transformers for language
  understanding.
\newblock {\em arXiv preprint arXiv:1810.04805}.

\bibitem[\protect\citename{Digman}1990]{digman1990personality}
John~M Digman.
\newblock 1990.
\newblock Personality structure: Emergence of the five-factor model.
\newblock {\em Annual review of psychology}, 41(1):417--440.

\bibitem[\protect\citename{Hearst \bgroup et al.\egroup
  }1998]{hearst1998support}
Marti~A. Hearst, Susan~T Dumais, Edgar Osuna, John Platt, and Bernhard
  Scholkopf.
\newblock 1998.
\newblock Support vector machines.
\newblock {\em IEEE Intelligent Systems and their applications}, 13(4):18--28.

\bibitem[\protect\citename{Kim \bgroup et al.\egroup }2002]{kim2002support}
Hyun-Chul Kim, Shaoning Pang, Hong-Mo Je, Daijin Kim, and Sung-Yang Bang.
\newblock 2002.
\newblock Support vector machine ensemble with bagging.
\newblock In {\em International Workshop on Support Vector Machines}, pages
  397--408. Springer.

\bibitem[\protect\citename{Liem \bgroup et al.\egroup
  }2018]{liem2018psychology}
Cynthia~CS Liem, Markus Langer, Andrew Demetriou, Annemarie~MF Hiemstra,
  Achmadnoer~Sukma Wicaksana, Marise~Ph Born, and Cornelius~J K{\"o}nig.
\newblock 2018.
\newblock Psychology meets machine learning: Interdisciplinary perspectives on
  algorithmic job candidate screening.
\newblock In {\em Explainable and Interpretable Models in Computer Vision and
  Machine Learning}, pages 197--253. Springer.

\bibitem[\protect\citename{Mairesse \bgroup et al.\egroup
  }2007]{mairesse2007using}
Fran{\c{c}}ois Mairesse, Marilyn~A Walker, Matthias~R Mehl, and Roger~K Moore.
\newblock 2007.
\newblock Using linguistic cues for the automatic recognition of personality in
  conversation and text.
\newblock {\em Journal of artificial intelligence research}, 30:457--500.

\bibitem[\protect\citename{Majumder \bgroup et al.\egroup
  }2017]{majumder2017deep}
Navonil Majumder, Soujanya Poria, Alexander Gelbukh, and Erik Cambria.
\newblock 2017.
\newblock Deep learning-based document modeling for personality detection from
  text.
\newblock {\em IEEE Intelligent Systems}, 32(2):74--79.

\bibitem[\protect\citename{Maria~Balmaceda \bgroup et al.\egroup
  }2014]{maria2014personality}
Jose Maria~Balmaceda, Silvia Schiaffino, and Daniela Godoy.
\newblock 2014.
\newblock How do personality traits affect communication among users in online
  social networks?
\newblock {\em Online Information Review}, 38(1):136--153.

\bibitem[\protect\citename{Mehta \bgroup et al.\egroup
  }2020]{mehta2020personality}
Yash Mehta, Navonil Majumder, Alexander Gelbukh, and Erik Cambria.
\newblock 2020.
\newblock Recent trends in deep learning based personality detection.
\newblock {\em Artificial Intelligence Review}, 53:2313–2339.

\bibitem[\protect\citename{Mikolov \bgroup et al.\egroup
  }2013]{mikolov2013efficient}
Tomas Mikolov, Kai Chen, Greg Corrado, and Jeffrey Dean.
\newblock 2013.
\newblock Efficient estimation of word representations in vector space.
\newblock {\em arXiv preprint arXiv:1301.3781}.

\bibitem[\protect\citename{Mohammad and Kiritchenko}2015]{mohammad2015using}
Saif~M Mohammad and Svetlana Kiritchenko.
\newblock 2015.
\newblock Using hashtags to capture fine emotion categories from tweets.
\newblock {\em Computational Intelligence}, 31(2):301--326.

\bibitem[\protect\citename{Pennebaker \bgroup et al.\egroup
  }2001]{pennebaker2001linguistic}
James~W Pennebaker, Martha~E Francis, and Roger~J Booth.
\newblock 2001.
\newblock Linguistic inquiry and word count: Liwc 2001.
\newblock {\em Mahway: Lawrence Erlbaum Associates}, 71(2001):2001.

\bibitem[\protect\citename{Pennington \bgroup et al.\egroup
  }2014]{pennington2014glove}
Jeffrey Pennington, Richard Socher, and Christopher Manning.
\newblock 2014.
\newblock Glove: Global vectors for word representation.
\newblock In {\em Proceedings of the 2014 conference on empirical methods in
  natural language processing (EMNLP)}, pages 1532--1543.

\bibitem[\protect\citename{Yin \bgroup et al.\egroup }2018]{yin2018network}
Han Yin, Yue Wang, Qian Li, Wei Xu, Ying Yu, and Tao Zhang.
\newblock 2018.
\newblock A network-enhanced prediction method for automobile purchase
  classification using deep learning.
\newblock In {\em PACIS}, page 111.

\end{thebibliography}

\end{document}